# Real-time Quasi-Optimal Trajectory Planning for Autonomous Underwater Docking

A.M. Yazdani, K. Sammut, A. Lammas, Y. Tang
Centre for Maritime Engineering, Control and Imaging
School of Computer Science, Engineering and Mathematics
Flinders University, Adelaide, SA 5042, Australia

*Abstract* - **In this paper, a real-time quasi-optimal trajectory planning scheme is employed to guide an autonomous underwater vehicle (AUV) safely into a funnel-shape stationary docking station. By taking advantage of the direct method of calculus of variation and inverse dynamics optimization, the proposed trajectory planner provides a computationally efficient framework for autonomous underwater docking in a 3D cluttered undersea environment. Vehicular constraints, such as constraints on AUV states and actuators; boundary conditions, including initial and final vehicle poses; and environmental constraints, for instance no-fly zones and current disturbances, are all modelled and considered in the problem formulation. The performance of the proposed planner algorithm is analyzed through simulation studies. To show the reliability and robustness of the method in dealing with uncertainty, Monte Carlo runs and statistical analysis are carried out. The results of the simulations indicate that the proposed planner is well suited for real-time implementation in dynamic and uncertain environment.**

*Keywords—quasi-optimal trajectory; AUV; underwater docking; uncertainty; IDVD*

## I. Introduction

In recent decades, AUVs have been increasingly employed for various underwater missions and explorations. Improvements in underwater sensor suites and embedded computer systems, have been a technology enabler for advancing AUV autonomy and is widening the scope of potential applications in which AUVs can be employed. One major factor, however, that still limits an AUV's mission endurance and thus its eligibility for long persistent autonomy, is its energy storage capacity which is constrained by the vehicle's geometric constraints and by budget limitations [1]. If a vehicle is incapable of carrying sufficient energy to complete its mission then the ability for the vehicle to recharge its batteries en-route should be considered.

Docking stations provide long term sustainability for AUVs by means of battery recharging and data communication units in such a way that the vehicle is able to upload the collected mission data and download new mission information. The funnel-shaped docking station is the most common structure and facilitates unidirectional approach for docking of the AUV, as shown in Fig.1.

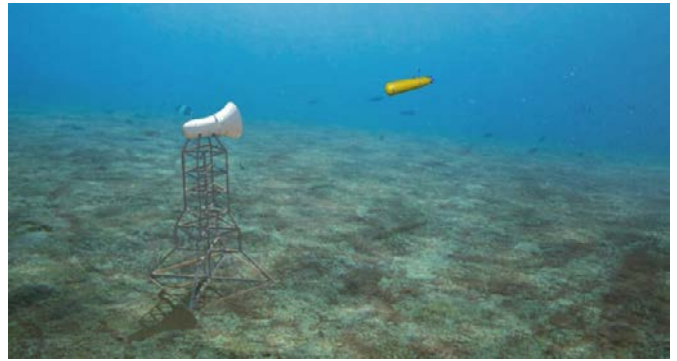

Fig.1. Underwater docking operation using a funnel-shaped dock

Several research studies have been conducted for underwater docking guidance of AUVs with funnel-shaped recovery stations [2], [3], [4]. Classic guidance strategies such as, pure pursuit guidance [5] and linear terminal guidance [6] have been typically used for guiding an AUV to its docking station. More recently, AI-based techniques such as fuzzy logic [5] and evolutionary algorithms [6] have also been used to address the same problem. In most of the docking literature, experiments have been conducted under controlled and roughly ideal conditions [7]. However, one of the most important considerations of the docking process is the ability to safely maneuver an AUV into the dock's funnel structure while allowing for all vehicular and environmental constraints. To this end, ocean currents that may cause discrepancies between the desired and actual AUV headings need to be modelled in the guidance block. The constraints on the kinematics and dynamics of the vehicle, as well as the controls, need to be addressed by the guidance module. In this respect, the planner algorithm is not only required to generate an appropriate path for the vehicle, but it must also entail the history of the vehicle's states such as components of the translational and angular velocities. Therefore, the problem of underwater docking can be mapped into a trajectory planning space.

The approach proposed in this paper employs the direct optimization method, referred to as Inverse Dynamics in the Virtual Domain (IDVD) [8], to provide a rapid 3D trajectory generation for the underwater docking problem. The main contribution of this work is to reformulate the docking problem to form a new and realistic perspective constrained with details of the vehicle's model, the geometric representation of the docking station and the local ocean current for providing a guidance framework. This guidance system, by means of a few parameters, can generate smooth computationally efficient trajectories suitable for real-time application. The generated trajectories are able to satisfy all boundary conditions and dynamic constraints when docking is performed within an actual ocean operating environment.

The organization of the paper is as follows. In Section II, the problem formulation is presented. Section III describes the IDVD method and its application to the trajectory planning problem. Numerical experiments and conclusions are then reported in Section IV and Section V, respectively.

## II. PROBLEM FORMULATION

This section first describes the motion model of the vehicle in the configuration space adopted for generating an admissible set of trajectories and thereafter presents the geometrical model of the docking station.

Three standard assumptions here are taken into account. First, the AUV is considered as a rigid body structure. Second, the Earth's rotation is ignored. The third assumption entails the linearized dynamic differential equation for translational and angular velocities. In the following, the kinematics of the AUV can be described as a set of ordinary differential equation as shown in (1) [9],

$$\begin{bmatrix} \dot{X} \\ \dot{Y} \\ \dot{Z} \end{bmatrix} = \begin{bmatrix} _b^n R \end{bmatrix} \begin{bmatrix} u \\ v \\ w \end{bmatrix} + \begin{bmatrix} u_c \\ v_c \\ 0 \end{bmatrix} \quad (1)$$

where $u$, $v$ and $w$ are surge, sway and heave velocities through the water in the body frame $\{b\}$, respectively. $\begin{bmatrix} _b^n R \end{bmatrix}$ is a rotation matrix that transforms the body frame $\{b\}$ into the NED frame $\{n\}$ using Euler angles, pitch $\theta$, yaw $\psi$ and roll $\varphi$ ($\varphi$ is ignored here), as expressed in (2).

$$_b^n R = \begin{bmatrix} \cos\psi\cos\theta & -\sin\psi & \cos\psi\sin\theta \\ \sin\psi\cos\theta & \cos\psi & \sin\psi\sin\theta \\ -\sin\theta & 0 & \cos\theta \end{bmatrix} \quad (2)$$

The symbols of $u_c$ and $v_c$ are the components of the 2D current velocity along the north and east direction, respectively, with respect to the ground in the NED frame.

The original conceptual model of the docking station as described in [10] is modified in this work to include a conical shape defining the boundaries of the permissible trajectories for successful entry into the dock. This more complete version is modelled using three major properties; dock position ($\sigma_{dock}$) and pose ($\psi_{dock}, \theta_{dock}$) in the NED frame together with the dock's geometric features illustrated in Fig.2. Table 1 specifies the dock's corresponding parameters in detail. These characteristics serve as the set of constraints for the trajectory generation.

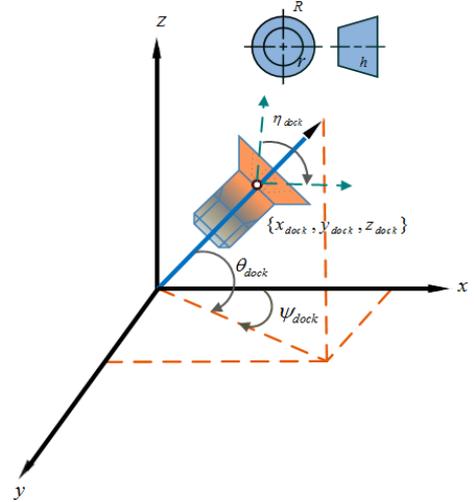

Fig.2. Geometric model of the docking station

TABLE I. SPECIFICATION OF THE DOCKING STATION

| Position | $\sigma_{dock} = \{x_{dock}, y_{dock}, z_{dock}\}$ | | | |
|---|---|---|---|---|
| Direction | *Horizontal* | | *Vertical* | |
| | $\psi_{dock}$ | | $\theta_{dock}$ | |
| | $h$ | $R$ | $r$ | $\eta_{dock}$ |
| **Geometric Features** | Length of the cone | Outer radius of the cone | Internal radius of the cone | Entrance angle of the cone |

The trajectory planning framework for guiding the AUV safely toward the docking station is thus expressed in the form of finding the set of admissible trajectories, defined in (3), that should satisfy the system of ordinary differential equations, typically defined by the:

- kinematics of vehicle as defined in (1),
- initial and final boundary conditions associated with initial and final pose of the vehicle , initial and final velocities, and initial and final accelerations as indicated in (4),
- constraints on states, controls and their derivatives, as defined in (5) and (6), respectively.

$$\Gamma(t) = [x(t), y(t), z(t), \psi(t), \theta(t), u(t), v(t), w(t)]^T \in S$$
$$S = \{\Gamma(t) \in \Gamma^8 \subset E^8\} \quad (3)$$
$$t \in [t_0, t_f]$$

s.t

$$\Gamma(t_0) = \Gamma_0, \Gamma(t_f) = \Gamma_f \quad (4)$$

$$\Gamma_{min} < \Gamma(t) < \Gamma_{max} \quad (5)$$

$$\xi(\Gamma, \dot{\Gamma}, t) \geq 0 \quad (6)$$

where $x(t), y(t), z(t)$ in (3) represent the history of position states in $x, y, z$ dimensions, respectively, in the NED frame. The environmental constraints in this paper are associated with the current disturbance, the no-fly zone, and the docking station specifications. The 2D current that causes drift between desired and actual heading of the vehicle, is employed with magnitude $|\vec{V}_C|$ and direction $\psi_c$ in the NED frame. It is essential to consider the discrepancy between desired and actual heading, particularly during the terminal phase of docking, to prevent collision with the docking structure. This issue applies to both horizontal and vertical approach angles ($\psi_{app}, \theta_{app}$). The difference between the actual heading angle (course angle) $\chi$ and dock direction in both horizontal and vertical plane, should be less than $\eta_{dock}/2$ (see Table 1) and the final position of the vehicle should be at $\sigma_{dock}$. Furthermore, the yaw and pitch angles are required to converge to $\psi_{dock}$ and $\theta_{dock}$ respectively, for successful entry into the dock [11]. Consequently, a constraint, given in (7), is imposed on the geometric path to prevent the AUV moving through the no-fly zone, denoted by $\Re$, during the docking operation.

$$[x(t), y(t), z(t)]^T \cap \Re = 0 \quad (7)$$

### III. IDVD Trajectory Planning Framework

The direct optimization method is a branch of optimal control theory which is utilized in rapid generation of optimal or sub-optimal solutions for complicated nonlinear systems. It converts the optimal control problems into a finite dimensional nonlinear programming problem (NLP) by using discretization and solving the problem for the states and controls based on the optimization routine [12], [13].

The basis of the IDVD method is to use the concept of the differential flatness theory [14], for performing the optimization in the output space by expressing the states and controls as functions of the output and its derivatives. Meanwhile, the optimization is operated in the virtual domain as opposed to the time domain, using an abstract argument $\tau$, allowing for fast prototyping of the optimal trajectories [8].

In general, IDVD method can be summarized in four steps below:

– **Step 1** Generate a reference function in virtual domain ($\tau$ domain) that is independent of time derivative constraints.
– **Step 2** Convert the reference trajectory back into time domain using speed factor ($\lambda$).
– **Step 3** Employ inverse dynamics to calculate states and controls.
– **Step 4** Operate optimization routine considering boundary condition, constraints and performance index.

For the proposed underwater docking problem, the reference functions are defined for the three spatial coordinates $x$, $y$, $z$, and for $\psi$. The AUV's path is parameterized in a form of a 7th order polynomial with trigonometric terms as shown in (8). This spatial reference function gives more flexibility for varying the curvature of the trajectory by providing higher order derivatives at the initial and end points [15].

$$x_i(\tau) = \sum_{k=0}^{5} a_{ik} \tau^k + b_{1i} \sin(\pi \tau) + b_{2i} \sin(2\pi \tau) \quad i = 1,2,3 \quad (8)$$

$$x_1(\tau) \equiv x(\tau) \quad , \quad x_2(\tau) \equiv y(\tau) \quad , \quad x_3(\tau) \equiv z(\tau)$$

All eight coefficients $a_{ik}$, $k = 0,...,5$ and $b_1, b_2$ together with the initial and final boundary conditions on states and their derivatives construct a linear matrix (9).

$$\begin{bmatrix} 1 & 0 & 0 & 0 & 0 & 0 & 0 & 0 \\ 0 & 1 & 0 & 0 & 0 & 0 & \pi & 2\pi \\ 0 & 0 & 2 & 0 & 0 & 0 & 0 & 0 \\ 0 & 0 & 0 & 6 & 0 & 0 & -\pi^3 & -8\pi^3 \\ 1 & 1 & 1 & 1 & 1 & 1 & 0 & 0 \\ 0 & 1 & 2 & 3 & 4 & 5 & -\pi & 2\pi \\ 0 & 0 & 2 & 6 & 12 & 20 & 0 & 0 \\ 0 & 0 & 0 & 6 & 24 & 60 & \pi^3 & -8\pi^3 \end{bmatrix} \begin{bmatrix} a_0 \\ a_1 \\ a_2 \\ a_3 \\ a_4 \\ a_5 \\ b_1 \\ b_2 \end{bmatrix} = \begin{bmatrix} x_0 \\ \dot{x}_0 \tau_f \\ \ddot{x}_0 \tau_f^2 \\ \dddot{x}_0 \tau_f^3 \\ x_f \\ \dot{x}_f \tau_f \\ \ddot{x}_f \tau_f^2 \\ \dddot{x}_f \tau_f^3 \end{bmatrix} \quad (9)$$

Mathematically speaking, the order of the polynomial ($N$) is dependent on the number of boundary conditions that must be satisfied. In this study, the initial and terminal states, first- and second-order derivatives are a group of constraints that must be met. The third-order derivatives, initial and final jerk, and $\tau_f$ are free variables that can be used for adjusting the trajectory. Hence, $N$ can be determined from (10), where

$d_0$, $d_f$ are the orders of the time derivatives of the initial and final boundary conditions.

$$N = d_0 + d_f + 1 \quad (10)$$

The reference function for $\psi$ is a fifth-order algebraic polynomial as defined in (11).

$$\psi(\tau) = \sum_{k=0}^{5} a_k \tau^k$$

$$\begin{bmatrix} 1 & 0 & 0 & 0 & 0 & 0 \\ 0 & 1 & 0 & 0 & 0 & 0 \\ 0 & 0 & 2 & 0 & 0 & 0 \\ 1 & 1 & 1 & 1 & 1 & 1 \\ 0 & 1 & 2 & 3 & 4 & 5 \\ 0 & 0 & 2 & 6 & 12 & 20 \end{bmatrix} \begin{bmatrix} a_0 \\ a_1 \\ a_2 \\ a_3 \\ a_4 \\ a_5 \end{bmatrix} = \begin{bmatrix} x_0 \\ x_0' \tau_f \\ x_0'' \tau_f^2 \\ x_f \\ x_f' \tau_f \\ x_f'' \tau_f^2 \end{bmatrix} \quad (11)$$

Once the spatial trajectories are formulated in the virtual domain, it is essential to transform them back into the time domain. This transformation is made possible by the so-called speed factor in (12) [8].

$$\lambda(\tau) = \frac{d\tau}{dt} \quad (12)$$

The speed factor facilitates a degree of flexibility for the speed profile to be varied along the same trajectory as described in (13).

$$V(\tau) = \lambda(\tau)\sqrt{x_1'(\tau) + x_2'(\tau) + x_3'(\tau)} \quad (13)$$

Referring to the control flow of IDVD, inverse dynamics is next employed to determine the states and control variables of the AUV necessary for trajectory planning purposes.

Considering the kinematic relation in (1), and the 2D current profile, an admissible assumption is made that there is a negligible difference between the flight path angle ($\gamma$) and the pitch angle ($\theta$). Therefore, taking advantage of inverse dynamics, the flight path angle is defined as (14).

$$\gamma = \theta = tan^{-1} \frac{-z'}{\sqrt{x'^2 + y'^2}} \quad (14)$$

Following this procedure, the north, east and down components ($V_{rn}, V_{re}, V_{rd}$) of the resultant ground speed, $\vec{V}_r$, in the geographical frame are calculated in (15) and, finally, the course angle, $\chi$, is expressed in (16).

$$V_{rn} = |\vec{V}| \cos\theta \cos\psi + |\vec{V}_c| \cos\psi_c$$
$$V_{re} = |\vec{V}| \cos\theta \sin\psi + |\vec{V}_c| \sin\psi_c \quad (15)$$
$$V_{rd} = |\vec{V}| \sin\theta$$

$$\chi = \tan^{-1} \frac{V_{re}}{V_{rn}} \quad (16)$$

The final step of the trajectory planning framework based on the IDVD is to run an optimization process to find an optimum reference trajectory which is subjected to the boundary conditions and certain constraints. For this purpose, and to measure the optimality of the desired trajectory in a quantitative manner, it is of course necessary to define a cost function. The cost function employed is a combination of the performance index and penalty function as described in (17).

$$Cost = (t_f - t_0) + \sum_{i=1}^{n} w_i f(V_i) \quad (17)$$
$$V = [z\ u\ v\ \theta\ \dot{\psi}\ \psi_{app}\ \theta_{app}\ \vartheta]^T$$

where $t_0, t_f$ are initial and final times, respectively (in this case $t_f$ is free), $w_i f(V_i)$ is a weighted violation function that respects the kinodynamic constraints of the AUV. These violations comprise: depth violation ($z$), to prevent the path from straying outside the vertical operating limits, surge ($u$) and sway ($v$) velocity violations, pitch ($\theta$) and yaw rate ($\dot{\psi}$) violations, horizontal and vertical approach angle ($\psi_{app}$ and $\theta_{app}$) violations as used for terminal phase of the docking operation, and finally, violation ($\vartheta$) specified to keep the path out of the no-fly zone during the docking operation. The AUV's kinematic and path constraints used for the optimization process are adopted from [16].

## IV. SIMULATION RESULTS

In this section, the simulation results produced by the IDVD algorithm for underwater docking of the AUV are presented. The operating field is modelled as a realistic underwater environment to examine the performance of the docking algorithm. A 2D constant current disturbance of magnitude $|\vec{V}_C| = 0.35(m/s)$ and direction of $\psi_c = 45°$ is employed. Three no-fly zones are modelled as three fixed spheres of different radii positioned between the initial position and the docking station. The docking station is located horizontally in the NED plane ($\theta_{dock} = 0°$) at position $x_{dock} = 150(m), y_{dock} = 75(m), z_{dock} = 10(m)$ and with horizontal direction. The geometric dimensions of the REMUS dock structure are assumed here [10].

For the purposes of this study, the optimization problem was performed on a desktop PC with an Intel i7 3.40 GHz quad-core processor using the built-in function *fminsearch* in MATLAB® R2014a.

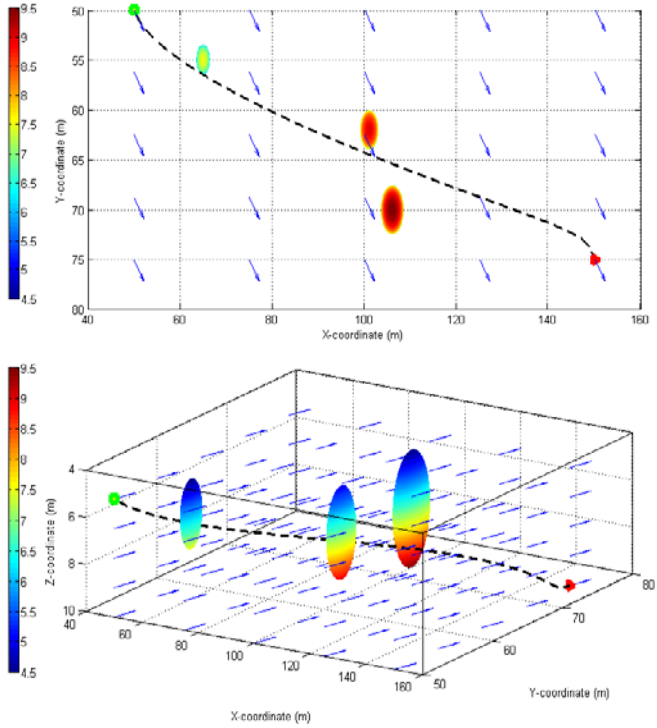

Fig.3. Docking approach: 2D overhead view and 3D geometric path

The vehicle starts its mission at initial location $x_{init}=50(m), y_{init}=50(m), z_{init}=5(m)$, horizontal direction $\psi_{init}=10°$, and vertical direction $\theta_{init}=0.5°$. Fig. 3 illustrates the generated 2D and 3D path that guides the AUV into the docking station. The corresponding surge and sway velocities along the path are shown in Fig. 4.

As can be seen from the graphs, the IDVD has generated acceptable smooth trajectories that meet the above constraints as denoted by the upper and lower (red) bounds. It is noteworthy to mention that the optimization takes less than a minute to process as it only uses a few parameters, making it highly suitable for real-time application.

Fig. 5 shows the changes in the AUV's heading and heading rate as it approaches the dock. Clearly, the heading trajectory converges to the required final direction that guarantees safe docking. The heading rate is smooth enough and realizable to pass on to a low level autopilot control module.

To validate the robustness of the proposed method, its performance is examined under uncertain operating conditions. A set of 100 Monte Carlo runs is conducted to statistically analyse the performance of the planning algorithm [17]. For this purpose, pseudorandom initial and final conditions equivalent to different initial and final poses of the AUV and the docking station in the problem space together with random current vectors are generated using both Gaussian and uniform distributions. Hence, three new experiments are conducted to evaluate the competence of IDVD in dealing with these variations. In the first experiment, the conditions are defined as: $x_{init}, y_{init}, x_{dock}, y_{dock} \sim N(0,10), z_{init}, z_{dock} \sim N(0,2)$ and $\psi_{init}, \psi_{dock} \sim N(0,45)$. In the second, the magnitude and direction of the current is varied according to $|\vec{V}_C| \sim N(0,0.3)$ and $\psi_c \sim N(0,90)$, respectively in addition to the first experiment's conditions. To make the underlying operating condition more challenging, in the third experiment the uncertainty is extended by using a uniform distribution of $\psi \sim U(0,360)$ for $\psi_{init}$, $\psi_c$ and $\psi_{dock}$. In this case, the variations completely cover all possible directions for the aforementioned parameters.

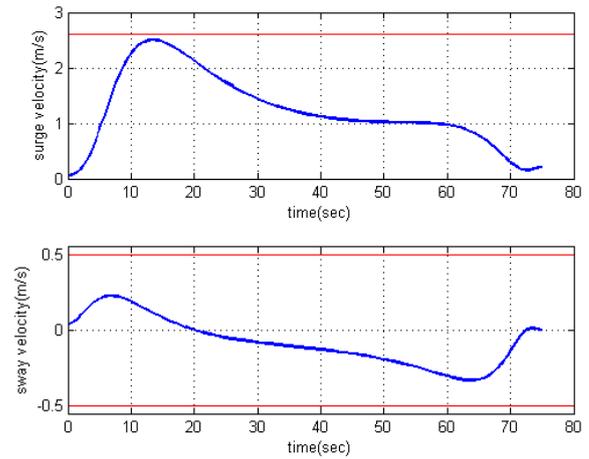

Fig.4. Surge and sway trajectories

To measure the performance optimality of the method in a quantitative manner, several performance metrics are utilized [18], [19]. Fig. 6 illustrates the statistical analysis of the proposed experiments against the results obtained without considering the variations in the mission scenarios (standard condition). Considering the index of the total flight time, path length and average speed, the results are quantitatively very similar in all experiments. More importantly, the violation percentage and the Root Mean Square Deviation (RMSD) performance metrics presented in Fig.6, reveal that the planner algorithm is robust to the variations of the vehicular parameters and environmental conditions and below the risk threshold with respect to the applied variations. The amount of error is not so significant that it can compromise the docking operation. Increasing the robustness of the method regarding the variation in spatial positions and directions of the docking station, can be facilitated by simultaneously updating the sensory information and re-planning strategy. By analysing the result of the Monte Carlo simulations, it is possible to be confident in the robustness and efficiency of the IDVD in dealing with uncertain undersea environments.

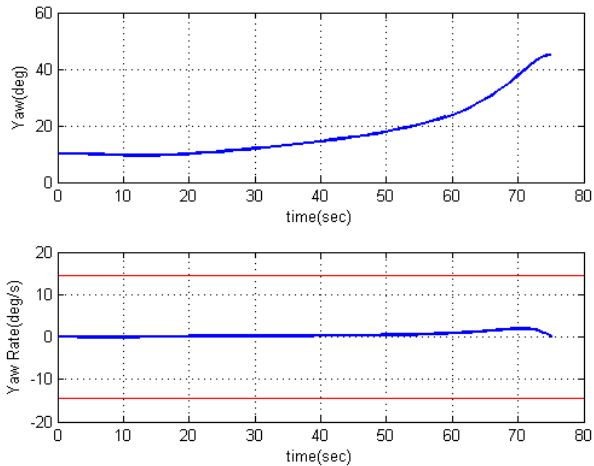

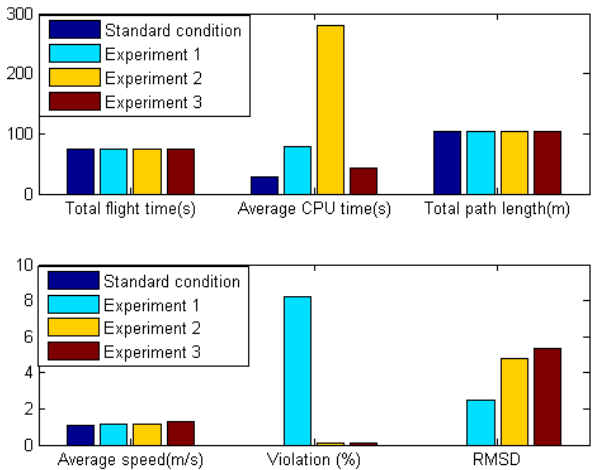

Fig.5. Heading and heading rates

Fig.6. Statistical analysis of the trajectory planner performance

## V. CONCLUSION

The problem of AUV underwater docking in the presence of current disturbances is addressed in this paper. By making use of the IDVD method, the docking problem is transformed into a trajectory planning case incorporating all the vehicular and environmental constraints. The proposed underwater docking trajectory planner framework is fast enough for real time application as only a few variables are used in the optimization process. The simulation results show that the method exhibits some sensitivity with respect to the uncertainty in the final desired pose. This can be problematic when the docking procedure is performed in a dynamic environment or with respect to a non-stationary docking station. Therefore, using re-planning is essential. Future work will concentrate on two aspects; firstly, further developing the current docking solution to produce a more dynamic version that can deal with partially known and unknown underwater environments, and secondly implementing the docking algorithm on a real vehicle for experimental validation.

REFERENCES

[1] K. Teo, E.An, J. Beaujean, "A Robust Fuzzy Autonomous Underwater Vehicle (AUV) Docking Approach for Unknown Current Disturbances", IEEE Journal of Oceanic Engineering, Vol. 37, No. 2, pp., 2012.
[2] D. Feezor, F. Yates Sorrell, R. Blankinship, G. Bellingham, "Autonomous Underwater Vehicle Homing/Docking via Electromagnetic Guidance", IEEE Journal of Oceanic Engineering, Vol. 26, No. 4, 2001.
[3] B.Allen, T.Austin, N.Forrester, R.Goldsborough, A. Kukulya, G. Packard, M. Purcell, R. Stokey, "Autonomous Docking Demonstrations with Enhanced REMUS Technology ", Proc.Oceans'2006, pp.1-6, 2006.
[4] B. Hobson, R. McEwen, J. Erickson, T. Hoover, L. McBride, F. Shane, and J. Bellingham, "The development and ocean testing of an AUV docking station for a 21" AUV", in Proc. Oceans'2007, pp.1-6, 2007.
[5] K. Teo, B. Goh, O. Chai, "Fuzzy Docking Guidance Using Augmented Navigation System on an AUV", IEEE Journal of Oceanic Engineering, Vol.40, Issue 2, pp. 349-361, 2015.
[6] Z. Yan, C. Deng, D. Chi, T.Chen, S.Hou , "Path Planning Method for UUV Homing and Docking in Movement Disorders Environment", The Scientific World Journal ,Vol.2014, pp.1-13, 2014.
[7] J. Y. Park, B. H. Jun, P. M. Lee and J. H. Oh., "Experiments on vision guided docking of an autonomous underwater vehicle using one camera", Ocean Engineering, Vol. 36, No. 1, pp. 48-61, 2009.
[8] A. Yakimenko, P. Kragelund "Real-Time Optimal Guidance and Obstacle Avoidance for UMVs", Autonomous Underwater Vehicles, Chap. 4, pp.67-98, 2011.
[9] I. Fossen, "Marine Control Systems Guidance, Navigation, and Control of Ships, Rigs and Underwater Vehicles", Marine Cybernetics, 2002.
[10] R.Stokey, B. Allen, T.Austin, R. Goldsborough, N.Forrester, M.Purcell, C.von Alt, "Enabling Technologies for REMUS Docking: An Integral Component of an Autonomous Ocean-Sampling Network", IEEE Journal of Oceanic Engineering, Vol.26 ,No.4,pp. 487-497, 2001.
[11] J. Park, B. Jun, P. Lee, Y. Lim, "Docking problem and guidance laws considering drift for an under-actuated AUV", Proc.Oceans'2011 MTS/IEEE, pp.1-7, 2011.
[12] Z. Yu, C. Jing, S. Lincheng, "Real-time trajectory planning for UCAV air-to-surface attack using inverse dynamics optimization method and receding horizon control " Chinese Journal of Aeronautics, Vol.26, No.4, pp.1038–1056 , 2013.
[13] R. Jamilnia, A. Naghash, "Optimal guidance based on receding horizon control and online trajectory optimization" Journal of Aerospace Engineering, Vol.26, No. 4, pp. 786-793, 2013.
[14] M. Fliess, J. Lévine, P. Martin, P. Rouchon, "Flatness and Defect of Nonlinear Systems: Introductory Theory and Examples", International Journal of Control, Vol.61, pp. 1327-1361, 1993.
[15] I. Cowling, O. Yakimenko, J. Whidborne, A. Cooke, "Direct Method Based Control System for an Autonomous Quadrotor", Journal of Intelligent and Robotic Systems, Vol.60, Issue 2, pp 285-316, 2010.
[16] S. Doherty, "Cross Body Thruster Control and Modelling of a Body of Revolution Autonomous Underwater Vehicle," Master's thesis, Naval Postgraduate School, 2011.
[17] Z. Zeng, A. Lammas, K. Sammut, F. He, Y. Tang, "Shell space decomposition based path planning for AUVs operating in a variable environment", Ocean Engineering, Vol. 91, pp. 181-195, 2014.
[18] A. Moeini, K. Jenab, M. Mohammadi, M. Foumani, "Fitting the three-parameter Weibull distribution with Cross Entropy", Applied Mathematical Modelling, Vol.37, pp. 6354–6363, 2013.
[19] A. Tanoto, V. Gomez, N. Mavridis, H. Li, U. Ruckert, S. Garrido, "Teletesting: Path Planning Experimentation and Benchmarking in the Teleworkbench", European Conference on Mobile Robots, pp. 343 348, 2013.